\documentclass[journal]{IEEEtran}
\usepackage{amsmath,amsfonts}
\usepackage{algorithm}
\usepackage{array}
\usepackage[caption=false,font=normalsize,labelfont=sf,textfont=sf]{subfig}
\usepackage{textcomp}
\usepackage{stfloats}
\usepackage{url}
\usepackage{verbatim}
\usepackage{cite}
\usepackage[colorlinks,linkcolor=red,citecolor=blue]{hyperref}

\ifCLASSINFOpdf
\usepackage[pdftex]{graphicx}
\DeclareGraphicsExtensions{.eps,.pdf,.tiff,.jpeg,.png}
\else
\usepackage[dvips]{graphicx}
\DeclareGraphicsExtensions{.eps}
\fi
\graphicspath{{pic/}{pic/authors}}

\usepackage{graphicx, algcompatible, epstopdf,epsfig,amssymb,multirow, enumerate,float, diagbox,makecell}
\usepackage[caption=false]{subfig}

\begin{document}

\title{A Memory-Augmented Multi-Task Collaborative Framework for Unsupervised Traffic Accident Detection in Driving Videos}

\author{{Rongqin Liang,~\IEEEmembership{Student Member,~IEEE,} 
Yuanman Li,~\IEEEmembership{Member,~IEEE,} 
Yingxin Yi,~\IEEEmembership{Student Member,~IEEE,} 
Jiantao Zhou,~\IEEEmembership{Senior Member,~IEEE,} and 
Xia Li,~\IEEEmembership{Member,~IEEE}}
\thanks{Rongqin Liang, Yuanman Li, Yingxin Yi and Xia Li are with Guangdong Key Laboratory of Intelligent Information Processing, College of Electronics and Information Engineering, Shenzhen University, Shenzhen 518060, China (email: 1810262064@email.szu.edu.cn; yuanmanli@szu.edu.cn; 2210433112@email.szu.edu.cn; lixia@szu.edu.cn. Corresponding author: Yuanman Li).

Jiantao Zhou is with the State Key Laboratory of Internet of Things for Smart City, and also with the Department of Computer and Information Science, University of Macau. e-mail: jtzhou@um.edu.mo.}
}



\maketitle

\begin{abstract}
Identifying traffic accidents in driving videos is crucial to ensuring the safety of autonomous driving and driver assistance systems. To address the potential danger caused by the long-tailed distribution of driving events, existing traffic accident detection (TAD) methods mainly rely on unsupervised learning. However, TAD is still challenging due to the rapid movement of cameras and dynamic scenes in driving scenarios.
Existing unsupervised TAD methods mainly rely on a single pretext task, \textit{i.e.}, an appearance-based or future object localization task, to detect accidents.
However, appearance-based approaches are easily disturbed by the rapid movement of the camera and changes in illumination, which significantly reduce the performance of traffic accident detection. 
Methods based on future object localization may fail to capture appearance changes in video frames, making it difficult to detect ego-involved accidents (\textit{e.g.}, out of control of the ego-vehicle). 
In this paper, we propose a novel memory-augmented multi-task collaborative framework (MAMTCF) for unsupervised traffic accident detection in driving videos. Different from previous approaches, our method can more accurately detect both ego-involved and non-ego accidents by simultaneously modeling appearance changes and object motions in video frames through the collaboration of optical flow reconstruction and future object localization tasks.
Further, we introduce a memory-augmented motion representation mechanism to fully explore the interrelation between different types of motion representations and exploit the high-level features of normal traffic patterns stored in memory to augment motion representations, thus enlarging the difference from anomalies.
Experimental results on recently published large-scale dataset demonstrate that our method achieves better performance compared to previous state-of-the-art approaches.
\end{abstract}

\begin{IEEEkeywords}
Traffic accident detection, Future object localization, Optical flow reconstruction, Transformer, Memory.
\end{IEEEkeywords}

\section{Introduction}
\IEEEPARstart{I}{N} recent years, autonomous driving technology \cite{9438625, 8715479} and advanced driver assistance systems \cite{7428841, 8809914} have achieved rapid development, bringing great convenience to human travel. At the same time, the everyday incidence of traffic accidents continually motivates efforts to increase the safety of driving systems in natural driving scenarios, especially in the detection of traffic accidents. Traffic accident detection (TAD) aims to detect abnormal traffic patterns in driving videos. Accurate TAD methods assist to reduce traffic accidents, increase road safety, shorten traffic recovery times and so on. 

Many researchers in computer vision have investigated the detection of anomalous events from dashboard-mounted cameras \cite{7564410, Liu_2018_CVPR, Liu_2021_ICCV, 9714213}. These methods can be mainly divided into supervised and unsupervised approaches. While supervised methods \cite{9714213, sun2022anomaly} have recently made great progress, the long-tailed distribution of driving events means that it may not be possible to collect all types of traffic accidents as training data, making it difficult to accurately detect traffic accidents outside the distribution of the dataset, potentially raising the risk of serious accidents. Therefore,
to overcome the difficulty of modeling all possible driving events, we focus on exploring an unsupervised TAD approach in driving videos.

With the rapid development of deep learning technology, unsupervised video anomaly detection (VAD) methods \cite{9828496, 8967556, 9712446, Gong_2019_ICCV} have significantly advanced in surveillance scenarios. However, in driving scenarios, high-speed moving vehicles cause videos with dynamically changing backgrounds, rendering these approaches not well-extended or even ineffective for driving scenarios \cite{9712446, 9733965}. 
Existing unsupervised TAD methods mainly model the normal traffic pattern by building a single pretext task, \textit{i.e.}, appearance-based \cite{8909850, 8943099, 9052726, Hasan_2016_CVPR} or future-object-localization-based \cite{8794474, Li2022, liang2021temporal} TAD methods, and treat observed events that deviate from the normal pattern as anomalies. 
Among them, appearance-based methods \cite{Liu_2018_CVPR, Hasan_2016_CVPR, 10.1007} focus on detecting differences between predicted or reconstructed video frames and natural frames, while future-object-localization-based methods \cite{8967556, 9712446} aim to detect anomalous motion by computing the variance of observed objects in predicted positions. Despite significant advances in these single-pretext-task-based methods, accurate detection of traffic accidents remains challenging. First, when the camera moves rapidly in the driving scene, appearance-based methods are easily disturbed by many factors such as dynamic background and illumination change, which may lead to misjudgment of TAD models. Second, although future-object-localization based methods avoid the difficulty of predicting whole frames, they generally fail to detect traffic accidents involving the ego-vehicle but not involving other objects (\textit{e.g.}, out of control of the ego-vehicle) due to their inability to capture appearance changes of video frames.
Therefore, how to accurately detect both ego-involved (\textit{i.e.}, traffic accidents involving the ego-vehicle) and non-ego (\textit{i.e.}, traffic accidents involving observed objects) accidents is very important for TAD in driving videos.
Besides, traffic accident detection is essentially about detecting outliers that distinguish them from normal traffic patterns. Therefore, it is important to model normal traffic patterns and improve the sensitivity to abnormal patterns for accurate traffic accident detection.
In this work, we argue that collaborating on optical flow reconstruction and future object localization tasks helps to more accurately detect both ego-involved and non-ego accidents. First, optical flow characterizes the appearance changes of video frames, which helps to detect ego-involved accidents. Second, accurate future object localization helps to detect abnormal object motion, which promotes the detection of non-ego accidents.
Additionally, optical flow in driving scenes also reflects the ego motion of the dashboard-mounted camera, which helps to better model the motion states of observed objects.
Furthermore, the motion state of observed objects reflects the local motion cues of video frames, which is potentially beneficial to modeling the appearance changes.

To fulfill the insights mentioned above, we propose a novel memory-augmented multi-task collaborative framework (MAMTCF) for unsupervised TAD in driving videos. First, we propose an unsupervised TAD framework that collaborates on optical flow reconstruction and future object localization tasks. Compared to existing TAD methods based on a single pretext task, our framework simultaneously modeling appearance changes and object motions in video frames, which helps detecting both ego-involved and non-ego accidents, achieving remarkable performance gains. In addition,
we propose a memory-augmented motion representation (MAMR) mechanism to model the interrelation between different types of motion representations, and utilize the high-level features of normal traffic patterns stored in memory to reconstruct motion representations. This enlarges the distinction from representations of abnormal traffic patterns and makes traffic accidents easier to detect.
Specifically, in the training phase, the MAMTCF is applied to train both the optical flow reconstruction and future object localization tasks. In the inference phase, we obtain an anomaly score for a driving video frame based on the reconstruction error of optical flow and the variance of predicted positions of observed objects. The main contributions of our work can be summarized as follows:
\begin{enumerate}
    \item We present a novel multi-task collaborative framework for unsupervised TAD. Compared to previous single-pretext-task-based TAD methods, 
    our framework models both appearance changes and object motions in video frames by collaborating on optical flow reconstruction and future object localization tasks. This collaboration promotes the detection of both ego-involved and non-ego accidents, greatly improving the detection of traffic accidents.
    \item We further propose a memory-augmented motion representation mechanism to fully explore the interrelation between different types of motion representations and reconstruct motion representations utilizing the high-level features of normal traffic patterns stored in memory. Such reconstructed motion representations help increase differences from anomalies, which benefits the detection of traffic accidents.
    \item The proposed framework achieves state-of-the-art performance on the recently published large-scale benchmark, providing a promising direction for unsupervised traffic accident detection in driving videos.
\end{enumerate}

The remainder of this paper is organized as follows. Section \ref{related_work} gives a brief review of related works. Section \ref{sec:proposed} details our proposed MAMTCF for traffic accident detection in driving videos. Extensive experimental results are presented in Section \ref{sec:experiment}, and we finally draw a conclusion in Section \ref{sec:conclusion}.
\section{Related Works} \label{related_work}
\subsection{Video Anomaly Detection (VAD) in Surveillance Videos}
Anomaly detection in surveillance videos aims to detect abnormal events occurring in the surveillance perspective. The main difference between VAD and TAD is that the background in surveillance videos is fixed, while the background in dashcam videos changes dynamically. 

Traditional VAD methods \cite{4407716, 4633642, 5206686, 5539872, Cheng_2015_CVPR, 6587741} mainly extract handcrafted features, followed by normality modeling to detect anomalies. For instance, Adam et al. \cite{4407716} designed multiple local, low-level feature (\textit{e.g.}, optical flow) monitors to detect abnormal events in the scene. Although traditional VAD methods demonstrate the importance of modeling normality, they rely on carefully handcrafted features and struggle to robustly handle various abnormal events.

With the rapid development of deep neural networks, researchers have recently proposed many deep-learning-based VAD methods. Among them, reconstruction-based methods and prediction-based methods are the two main paradigms of VAD methods. Reconstruction-based methods \cite{8019325, Luo_2017_ICCV, 8296547, FAN2020102920, 9410375, 9701300} typically train a generative model to reconstruct normal data, expecting the model to exhibit large reconstruction errors for abnormal data. For example, ConvLSTM-AE \cite{8019325} integrated a convolutional neural network (ConvNet) and a convolutional long short-term memory network (ConvLSTM) with an autoencoder to learn the regularity of appearance and motion at ordinary moments. Ravanbakhsh et al. \cite{8296547} employed generative adversarial networks (GANs) to learn an internal representation of scene normality to detect anomalies through reconstruction errors in appearance and motion representations. However, due to the potentially generalization capability of generative models, they can even reconstruct abnormal data, which can lead to the detection of some abnormal events being missed. To alleviate this problem, some researchers have proposed prediction-based VAD methods \cite{Nguyen_2019_ICCV, 8909850, 8943099, 9052726, 10.1145/3123266.3123451, 10.1145/3343031.3350899, 9645572}, which mainly rely on prediction errors to evaluate anomalies. For instance, Conv-VRNN \cite{8909850} introduced a sequence generation model based on Variational Autoencoder (CVAE) for future frame prediction with ConvLSTM. DMMNet \cite{9052726} provided a flexible masking network for motion and appearance fusion on video frame prediction. However, prediction-based methods may not be robust to noise in real surveillance videos, which can lead to a rapid reduction in detection performance. Additionally, some approaches \cite{TANG2020123, Ristea_2022_CVPR, Liu_2021_ICCV} attempt to combine the two paradigms of reconstruction and prediction. For instance, SSPCAB \cite{Ristea_2022_CVPR} integrated reconstruction-based functionality into a self-supervised predictive architecture building block. Although the aforementioned methods have achieved promising performance in the VAD task for surveillance videos, they are difficult to directly apply to traffic accident detection in a driving scenario. This is because the front and background of the video change dynamically due to the rapidly moving dashboard-mounted camera.  

\subsection{Traffic Accident Detection (TAD) in Dashcam Videos}
Traditional traffic accident detection methods \cite{7564410, YUAN2018202} mainly extract handcrafted features and classify them using a Bayesian model. Among them, Yuan et al. \cite{7564410} measured the abnormality of motion orientation and magnitude and fused the measurements using a Bayesian model to obtain a detection result. However, these methods are computationally complex, sensitive to handcrafted features, and lack robustness when applied to various traffic accidents.

With the advances in deep learning in computer vision, deep-learning-based traffic accident detection \cite{Liu_2018_CVPR, 8967556, 9712446, 9733965, 9714213, sun2022anomaly} has attracted the attention of researchers. Existing TAD methods mainly detect traffic accidents in an unsupervised manner through a single pretext task, \textit{i.e.}, appearance-based or future-object-localization-based TAD methods. Among them, appearance-based approaches \cite{Liu_2018_CVPR, 10.1007} focus on detecting the difference between the predicted or reconstructed frame and the natural frame. For instance, Liu et al. \cite{Liu_2018_CVPR} introduced appearance and motion constraints to facilitate the prediction of future frames for normal events and thus help identify anomalous events that do not conform to expectations. Another part of the works \cite{8967556, 9712446} applied future object localization to detect abnormal events. For instance, Yao et al. \cite{9712446} predict the future locations of objects over a short horizon in a driving scenario and then monitor prediction accuracy and consistency metrics as evidence of anomalies. Though previous single-pretext-task-based methods have achieved promising performance, these approaches still have inherent limitations, \textit{e.g.}, appearance-based methods are prone to greatly reduce performance due to the rapid movement of dashboard-mounted cameras, while methods based on future object localization may fail to capture changes in appearance, which may lead to failure to detect ego-involved accidents. Recently, Fang et al. \cite{9733965} attempted to collaborate frame prediction and future object localization tasks to absorb the merits of them. They proposed the SSC-TAD framework to detect traffic accidents by analyzing the inconsistency of video frames, object locations, and scene spatial relationship structures between different frames of driving videos. Different from SSC-TAD, our framework collaborates on the tasks of optical flow reconstruction and future object localization. The proposed MAMR mechanism not only fully explores the inherent interrelation between different types of motion representations but is also more sensitive to abnormal patterns.
\begin{figure*}[!t]
	\centering
	\includegraphics[width=0.79\linewidth]{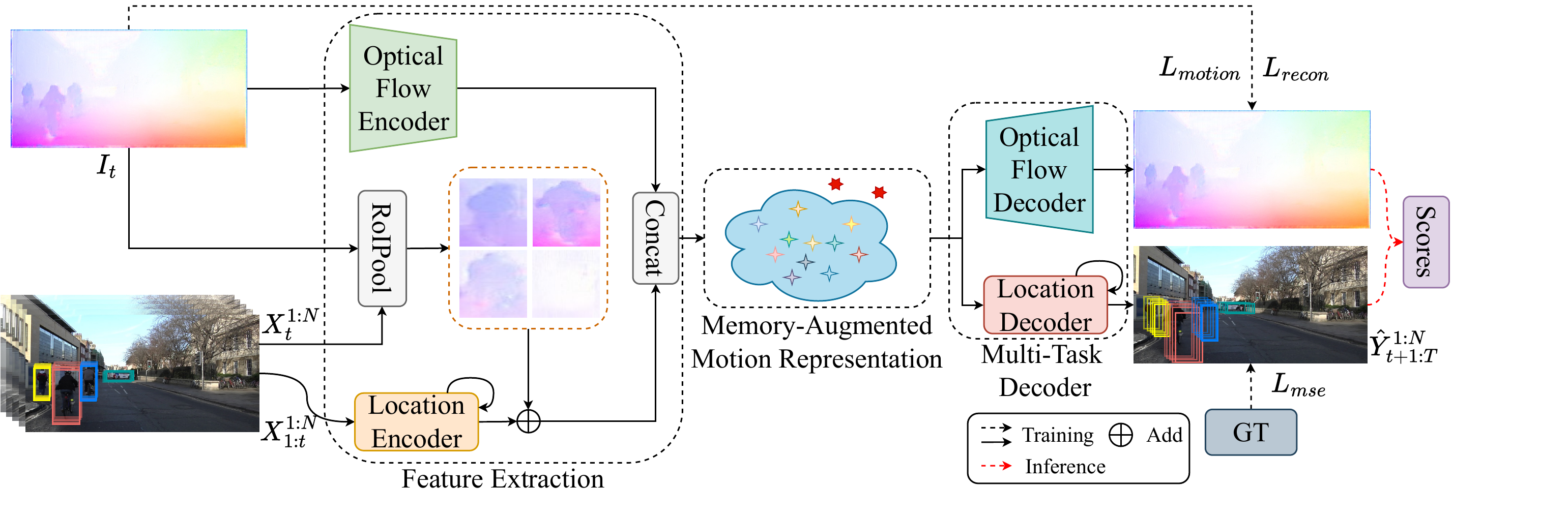}
	\caption{The framework of our MAMTCF algorithm. MAMTCF primarily consists of a feature extraction module, a memory-augmented motion representation mechanism, and a multi-task decoder. 1) First, the optical flow of driving video frames and the bounding boxes of observed objects in the scenes are encoded into different types of motion representations through the feature extraction module; 2) then, a memory-augmented motion representation (MAMR) mechanism collaborates on different types of motion representations and outputs the memory-augmented motion representations; 3) the memory-augmented motion representations are passed through the multi-task decoder to reconstruct optical flow and predict future bounding boxes of objects, respectively; 4) finally, the anomaly score is obtained by fusing the reconstruction error of the optical flow and variance of predicted bounding boxes.}
	\label{fig:framework}
\end{figure*}

Besides, some researchers have recently proposed supervised methods \cite{9714213, sun2022anomaly} for TAD in driving videos. For instance, Zhou et al. \cite{9714213} proposed a coarse-to-fine supervised TAD method based on spatio-temporal feature encoding with a multi-layer neural network. Sun et al. \cite{sun2022anomaly} proposed a traffic anomaly detection method based on cross-domain few-shot learning.
However, since driving events follow a long-tailed distribution, it may not be possible to collect all types of traffic accidents. This may increase the risk of the supervised TAD approach to cope with unknown events. Therefore, in our work, we focus on exploring an unsupervised TAD approach to avoid collecting all types of traffic incidents as training data.
\subsection{Memory Networks}
Memory modules in neural networks \cite{a27, wu-etal-2022-memformer} have recently gained much attention as a type of read-write global memory. For instance, Memformer \cite{wu-etal-2022-memformer} utilizes an external dynamic memory to encode and retrieve past information for efficient sequence modeling. More recently, some works \cite{Park_2020_CVPR, Liu_2021_ICCV, Gong_2019_ICCV} have applied memory networks to the VAD task in surveillance videos. For example, Park et al. \cite{Park_2020_CVPR} proposed using a memory module with an update scheme, where items in the memory record the normal patterns of the training data. $HF^2$-VAD \cite{Liu_2021_ICCV} introduces a multi-level memory module in an autoencoder with skip connections to memorize normal patterns for optical flow reconstruction, so that abnormal events can be sensitively identified through reconstruction errors. In this work, we make the first attempt to use a memory network to memorize normal traffic patterns for traffic accident detection in driving videos.
\section{The Proposed Approach: MAMTCF} \label{sec:proposed}
The overall framework of MAMTCF model is illustrated in Fig. \ref{fig:framework}. It primarily consists of three components:
1) a feature extraction module to extract different types of motion representations;
2) a memory-augmented motion representation (MAMR) mechanism to collaborate on different types of motion representations and output the memory-augmented motion representations; 
3) a multi-task decoder to reconstruct optical flow and predict future bounding boxes of objects. Note that the whole framework is trained on normal data only. In the inference phase, both the reconstruction and prediction errors are used for traffic accident detection.

In the following sections, we introduce the feature extraction module first, then the MAMR mechanism, followed by the multi-task decoder, and finally show how to use our model for traffic accident detection in driving videos.
\subsection{Feature Extraction}
Modeling changes in appearance of video frames as well as the motion states of observed objects in the scene is beneficial for detecting both ego-involved and non-ego traffic accidents. In our work,
we first apply the feature extraction module to extract different types of motion representations. As shown in Fig. \ref{fig:framework}, we primarily extract representations of two types of inputs, \textit{i.e.}, the optical flow of video frames and the bounding boxes of objects in observed scenes.

Practically, the optical flow of video frames not only contains the appearance changes of the scene but also reflects the ego motion of the ego-vehicle. To capture both the appearance changes and the ego motion simultaneously, we define the representation of optical flow from frame $t$ to $t + 1$ as the \textbf{\textit{global motion}} at time step $t$, which can be written as:
\begin{align}
F_t=\phi \left( I_t;\ \varPhi _f \right),
\end{align}
where $\varPhi _f$ is the parameter of the optical flow encoder $\phi(\cdot)$, $I_t$ denotes the optical flow from frame $t$ to $t + 1$, which is obtained by pre-trained FlowNet 2.0 \cite{Ilg_2017_CVPR}. Without loss of generality, we adopt the encoder in $HF^2$-VAD \cite{Liu_2021_ICCV} as the optical flow encoder $\phi(\cdot)$ in our experiments.

Moreover, the bounding boxes of observed objects in the scene reflect the motion states of the objects. Thus, we define the representation of bounding boxes of objects in observed scenes as the \textbf{\textit{object motion}}, which can be formulated as:
\begin{align}
    X_t=\varphi \left( X_{1:t}^{1:N};\ \varPhi _x \right),
\end{align}
where $\varPhi _x$ represents the parameter of the location encoder $\varphi(\cdot)$, $X_{1:t}^{1:N}$ denotes bounding boxes of $N$ objects in observed $ t$ scenes, which is obtained by pre-trained Mask-RCNN \cite{He_2017_ICCV}. In our experiments, the location encoder consists of a Fully Connected layer (FC) followed by a Gated Recurrent Neural network (GRU). 
In addition, to better perceive the motion states of objects, we further utilize the optical flow of objects to encode their current motion features. Specifically, similar to \cite{8967556, 9733965}, we extract the motion features of objects from the precomputed optical flow field using a region-of-interest pooling (RoIPool) operation with bilinear interpolation. 
The \textit{object motion} $X_t$ can be updated as follows:
\begin{align}
    X_t:=X_t+MLP\left( RoIPool\left( I_t,\ X_{t}^{1:N} \right); \varPhi _m \right),
\end{align}
where $\varPhi _m$ is the parameter of the Multilayer Perceptron (MLP). 
After obtaining the \textit{global motion} $F_t$ and the \textit{object motion} $X_t$, our model can perceive both the appearance changes of video frames and the motion states of objects in the scene, enabling our method to detect both ego-involved and non-ego traffic accidents.
\subsection{Memory-Augmented Motion Representation (MAMR) Mechanism for Collaborating Multi-Task}
Collaborating on different types of motion representations helps to detect different types of traffic incidents, \textit{i.e.}, \textit{global motion} focuses on appearance changes of video frames to detect ego-involved accidents, while \textit{object motion} emphasizes the motion states of the observed objects to detect non-ego accidents. 
Intuitively, the simplest way to fuse these two types of motion representations is to concatenate them and decode them for different pretext tasks. However, on the one hand, \textit{global motion} and \textit{object motion} have an inherent interrelation. The \textit{global motion} reflects the ego motion of the dashboard-mounted camera, which helps to better model the \textit{object motion} of the observed object and is beneficial to the detection of non-ego accidents. Correspondingly, \textit{object motion} reflects the local motion cues of the video frames, which potentially contribute to characterize the \textit{global motion} of video frames and thus promote the detection of ego-involved accidents. On the other hand, traffic accident detection is essentially about detecting outliers to distinguish them from normal traffic patterns. However, the potential generalization ability of the autoencoder leads to the possibility that it may learn shared patterns with abnormal traffic patterns \cite{Gong_2019_ICCV}, thereby blurring the distinctions between normal and abnormal motion representations. Consequently, this could lead to a failure to detect traffic accidents. Therefore, simply concatenating different types of motion representations is not a good strategy. This is also verified by experimental results in Table \ref{ablation}. 

In our work, we specially design a memory-augmented motion representation mechanism to collaborate different types of motion representations. As shown in Fig. \ref{fig:memory}, our MAMR mechanism mainly composes an inter-motion layer and a memory-augmented motion layer. The former models the interrelation between different types of motion representations while the latter reconstructs motion representations using normal traffic patterns stored in memory, thereby increasing differences from anomalies.
\subsubsection{Inter-motion layer}
\begin{figure*}[!t]
	\centering
	\includegraphics[width=0.70\linewidth]{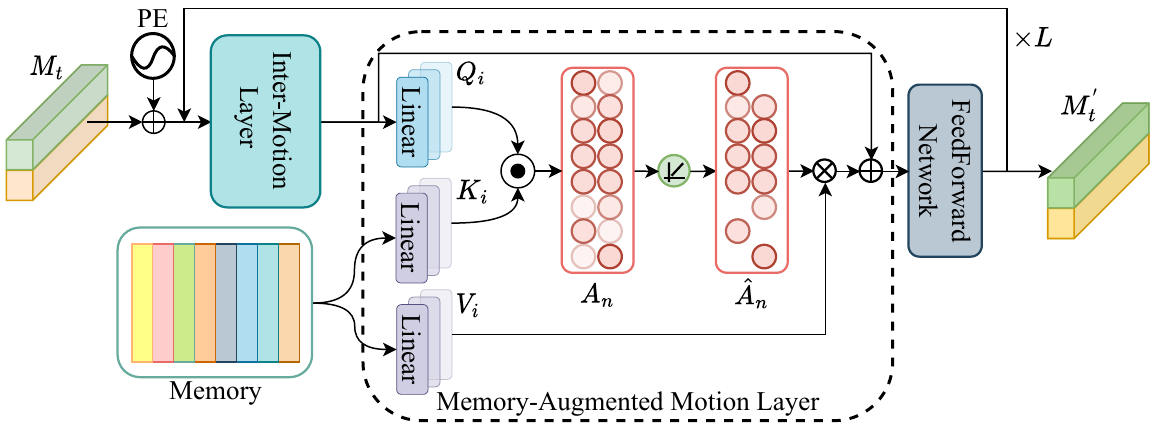}
	\caption{Illustration of the propose MAMR mechanism. The MAMR mechanism mainly consists of a inter-motion layer, memory-augmented motion layer, and feedforward network. The inter-motion layer explores the interrelation between \textit{global motion} and \textit{object motion}, while the memory-augmented motion layer retrieves high-level features of normal traffic patterns stored in memory and reconstructs motion representations with normal traffic patterns.}
	\label{fig:memory}
\end{figure*}

To model the interrelation between \textit{global motion} and \textit{object motion}, we first design a self-attention-based inter-motion layer for better modeling appearance changes and motion states of observed objects in video frames.
Specifically, as shown in Fig. \ref{fig:memory}, we first add a positional encoding to \textit{global motion} and \textit{object motion} to encode the relative relation of them, which can be formulated as:
\begin{equation}
    \begin{aligned}
    M_t = Concat(F_t, X_t),\\
    M_t := M_t + PE(M_t),\\
    \end{aligned}
\end{equation}
where $M_t$ is the concatenating representation of \textit{global motion} $F_t$ and \textit{object motion} $X_t$, $PE(\cdot)$ denotes the widely used hard-coded position embedding strategy \cite{vaswani2017NIPS}.
Then, we apply a self attention mechanism to model the interrelation between \textit{global motion} and \textit{object motion}, which can be formulated as:
\begin{equation}
    \begin{aligned}
    Q_m,K_m,V_m=M_tW_Q,M_tW_K,M_tW_V, \\
    A_t=MHAttn\left( Q_m,K_m \right), \\
    H_t=M_t + LN(Softmax \left( A_t \right) V_m),
    \end{aligned}
\end{equation}
where $W_Q, W_K$, and $W_V$ are the parameters corresponding to the Query $Q_m$, Key $K_m$, and Value $V_m$ of $M_t$, $MHAttn(\cdot)$ denotes the multi-head self-attention, $LN(\cdot)$ is the layer normalization, and $H_t$ models the interrelation between \textit{global motion} and \textit{object motion}.

\subsubsection{Memory-augmented motion layer}
Traffic accident detection is essentially about detecting outliers to distinguish them from normal traffic patterns. 
Intuitively, after modeling the interrelation between \textit{global motion} and \textit{object motion}, we can directly decode motion representations for multi-task. However, 
the autoencoder may learn some common features between normal and abnormal traffic patterns \cite{Gong_2019_ICCV}, thereby reducing the distinctions between normal and abnormal motion representations, which could result in missed detection of traffic accidents.
Therefore, 
we further propose a cross-attention-based memory-augmented motion layer. By utilizing the high-level features of normal traffic patterns stored in memory to reconstruct the motion representations, we can augment the modeling of normal traffic patterns and improve sensitivity to abnormal patterns.

Specifically, as shown in Fig. \ref{fig:memory}, the memory is designed as a matrix $\mathcal{M}\in \mathbb{R}^{M \times C}$ that contains $M$ slots of high-level features of normal traffic patterns with a fixed dimension $C$. Note that the memory $\mathcal{M}$ is randomly initialized at the beginning of training. 
For each input $H_t$, the memory-augmented motion layer needs to read the memory to retrieve relevant high-level features of normal traffic patterns. We leverage the cross-attention mechanism to achieve this function:
\begin{equation}
    \begin{aligned}
    Q_h,K_n,V_n=H_tW_Q^{'},\mathcal{M}W_K^{'},\mathcal{M}W_V^{'}, \\
    A_n=MHAttn\left( Q_h,K_n \right), \\
    H_n=H_t + LN(Hard Shrinkage \left( A_n \right) V_n),
    \end{aligned}
\end{equation}
where memory slot vectors are projected into Keys $K_n$ and Values $V_n$ with parameters $W_K^{'}$ and $W_V^{'}$, and the $H_t$ is projected into the Query $Q_h$ with parameter $W_Q^{'}$. $H_n$ denotes the reconstructed motion representations. Note that, to make the stored high-level features of normal traffic patterns more representative, we utilize the hard shrinkage operation in \cite{Gong_2019_ICCV} to promote the sparsity of the memory $\mathcal{M}$:
\begin{align}
    \hat{a}_i=h\left( a_i;\lambda \right) =\frac{ReLU\left( a_i-\lambda \right) \cdot a_i}{|a_i-\lambda |+\varepsilon},
\end{align}
where $\varepsilon$ is a very small positive scalar, $a_i\in A_n$, $i\in(1,M)$, $\lambda$ denotes the shrinkage threshold, and $ReLU$ is the ReLU activation. The hard shrinkage operation encourages the model to reconstruct $H_t$ with fewer but more relevant memory items, thus prompting the learning of high-level features that are more representative of normal traffic patterns in the memory. In addition, we minimize a sparsity regularizer on $\hat{A}_n$ during training to promote the sparsity of the memory, similar to \cite{Gong_2019_ICCV}:
\begin{align}
    L_s=\sum_{i=1}^M{-\hat{a}_i\cdot \log \left( \hat{a}_i \right)},
\end{align}
where $\hat{a}_i\in \hat{A}_n$ and $i\in(1,M)$.

Finally, we apply a feedforward network after the memory-augmented motion layer, which can be formulated as:
\begin{align}
    M_{t}^{'}=H_n+LN\left( FFN\left( H_n;\varTheta _f \right) \right),
\end{align}
where $FFN(\cdot)$ is a block of two fully connected layers, $\varTheta _f$ denotes the parameters of $FFN(\cdot)$, and $M_{t}^{'}$ is the memory-augmented motion representation.
We can observe that our MAMR mechanism not only models the interrelation between different types of motion representations, but also utilizes the high-level features of normal traffic patterns to reconstruct motion representations, enhancing the sensitivity of our framework to abnormal traffic accidents.
\subsection{Multi-Task Decoder for Traffic Accident Detection}
After obtaining the memory-augmented motion representation, we can easily decode it for multiple tasks, \textit{i.e.}, optical flow reconstruction and future object localization. During the inference phase, unsupervised TAD is implemented based on the reconstruction error and prediction bias of multi-tasks. 
\subsubsection{Optical flow reconstruction}
As shown in Fig. \ref{fig:framework}, we utilize the multi-task decoder to reconstruct the optical flow of video frames and predict the future bounding boxes of observed objects in the scene. Specifically, the motion representation $M_{t}^{'}$ is passed through the optical flow decoder to reconstruct the optical flow, which can be formulated as:
\begin{equation}
    \begin{aligned}
    F_t^{'}, X_t^{'} =Split(M_t^{'}), \\
    I_{t}^{'}=\phi _{dec}(F_t) ;\varTheta _f),
    \end{aligned}
\end{equation}
where $Split(\cdot)$ operation represents the partition of $M_t^{'}$ by channel, and $F_t^{'}$, $X_t^{'}$ denote the memory-augmented \textit{global motion} and \textit{object motion}, respectively. $\varTheta _f$ is the parameters of the optical flow decoder $\phi _{dec}(\cdot)$, and $I_{t}^{'}$ denotes the reconstructed optical flow. In our experiments, the decoder in $HF^2$-VAD \cite{Liu_2021_ICCV} is utilized as the optical flow decoder. Note that, to better reconstruct the optical flow, we replace the convolutional layer of the penultimate layer in the decoder with the SSPCAB module \cite{Ristea_2022_CVPR}. This module integrates the reconstruction-based functionality into a self-supervised prediction architecture building block, as detailed in \cite{Ristea_2022_CVPR}.

In fact, in driving scenarios, the moving velocity is one of the important factors leading to traffic accidents. Therefore, in our work, we not only supervise the consistency of the optical flow but also emphasize the consistency of the reconstructed motion. The loss function for the reconstructed optical flow is as follows:
\begin{equation}
    \begin{aligned}
        L_f&=L_{motion}+L_{recon} \\
       &=\sqrt{\left( I_x-I_{x}^{'} \right) ^2+\left( I_y-I_{y}^{'} \right) ^2}+|I_t-I_{t}^{'}|,
    \end{aligned} \label{loss_flow}
\end{equation}
where $(I_x, I_y)=I_t$ denotes the the offset of the image in $x$ and $y$ directions. The first term $L_{motion}$ of formula \ref{loss_flow} emphasizes motion consistency, while the last term $L_{recon}$ supervises the reconstructed optical flow.
\subsubsection{Future object localization}
For the future object localization task, we utilize the location decoder to recurrently decode the motion representation $M_{t}^{'}$ to future bounding boxes of objects, which can be written as:
\begin{equation}
    \begin{aligned}
        h_t = g(M_t^{'}; \xi _m), \\
        h_{t+1}=GRU\left( e_{t},h_t;\xi _g \right),\\
        e_{t+1}=MLP(h_{t+1}; \xi _e),\\
        \hat{Y}_{t+1}^{1:N}=MLP\left( h_{t+1};\xi _y \right),
    \end{aligned}
\end{equation}
where $\xi _*$ is the learnable parameters, and the projection head $g(\cdot)$ consist of a linear layer followed by a ReLU activation layer. In addition, $e_0$ is initialized with zeros. $\hat{Y}_{t+1}^{1:N}$ denotes the predicted bounding boxes of objects at timestep $t+1$, where $t$ ranges from $t$ to $T-1$.

Further, the loss function for the predicted bounding box can be defined as:
\begin{align}
    L_{mse}=\sum_{i=t+1}^T{\frac{1}{N}\sum_{n=1}^N{d\left( Y_{i}^{n},\hat{Y}_{i}^{n} \right)}},
\end{align}
where $d(\cdot)$ calculates the Euclidean distance. Combined with the loss function of the reconstructed optical flow, our final loss is formulated as:
\begin{equation}
    L_{total} = \lambda _1 L_f + \lambda _2 L_{mse} + \lambda _3 L_{s},
    \label{loss_total}
\end{equation}
where $\lambda _1$, $\lambda _2$, and $\lambda _3$ are the coefficients of different losses.
\subsubsection{Traffic accident detection}
In this section, we present how our multi-task collaborative framework detects traffic accidents in driving videos during the inference phase. 

Specifically, we fuse the traffic anomaly score based on motion consistency with the variance of the predicted bounding box. We calculate the reconstruction error of the motion and the variance of the predicted bounding box separately as follows:
\begin{equation}
    \begin{aligned}
        &S_e=\sqrt{( I_x-I_{x}^{'}) ^2+( I_y-I_{y}^{'} ) ^2}, \\
        S_l=&\underset{\{ 1:N \}}{\max}( \underset{\{ bbox \}}{\text{mean}}( ( STD( [ |Y_{t,t-j}^{1:N}-\hat{Y}_{t,t-j}^{1:N}| ] _{j=1}^{j=\delta} ) ) ) )
    \end{aligned}
    \label{formula}
\end{equation}
where $S_e$ denotes the motion reconstruction error, and $S_l$ represents the variance of $\delta$ bounding boxes at time $t$ predicted from time $t-1, t-2, ..., t-\delta$. $S_e$ focuses on appearance changes of video frames, which helps to detect ego-involved traffic accidents. $S_l$ emphasizes the motion behavior of objects in the scene, which is beneficial to detect non-ego accidents.
Note that we calculate the variance by $STD(\cdot)$ for the top-left and bottom-right coordinates of the bounding box (\textit{i.e.}, $bbox={x_{min},y_{min},x_{min},y_{min}}$) and take the average as the prediction error for that object at time $t$. Given that objects in a scene with an accident have a relatively large corresponding prediction error, therefore, we take the maximum prediction error in the scene as the anomaly score of the driving video at time $t$.

Furthermore, we fuse the motion reconstruction error $S_e$ with the variance of the bounding box $S_l$ to obtain the final traffic anomaly score $S_f$, which can be formulated as:
\begin{align}
    S_f=Norm\left( \alpha Norm\left( S_e \right) + (1-\alpha) Norm\left( S_l \right) \right),
\end{align}
where $Norm(\cdot)$ denotes the max-min normalization, and $\alpha \in [0, 1]$ represents the fusion coefficient.
\section{Experiments and Discussions}  \label{sec:experiment}
In this section, we evaluate the performance of our proposed method, which is performed on a platform with one NVIDIA 3090 GPU. All experiments were implemented using the PyTorch framework. Our source code and trained models will be publicly available upon acceptance. 

\subsection{Implementation Details}
In the experiment, we resize the extracted optical flow of video frames to $64 \times 64$. Besides, we observe $5$ frames of previous bounding boxes and predict $10$ frames of future bounding boxes. The dimension of \textit{global motion} and \textit{object motion} is set to $512$, and the size of the ROI pooling operation is set to $5 \times 5$. We set the layer of the MAMR mechanism to 3 (\textit{i.e.}, $L=3$), and empirically set the head number of self-attention and cross-attention to $8$. The slot number $M$ is empirically set as 1000, and $\lambda$ in the hard shrinkage operation is set to $3/M$. In the training phase, we empirically set the coefficients $\lambda _1$ and $\lambda _2$ of the final loss to be $1.0$ and $1.0$, respectively. Following prior works \cite{Liu_2021_ICCV, Ristea_2022_CVPR}, we set the coefficient $\lambda _3$ to $0.0002$ in practice. We optimize the loss function (\ref{loss_total}) using Adam algorithm \cite{kingma2014adam} with a batch size of 128, learning rate of 1e-4, betas of 0.9 and 0.999, weight decay of 5e-4, and train our framework for 100 epochs. During inference, we experimentally set the fusion coefficient $\alpha$ to 0.4.
\begin{table}[!t]
\renewcommand{\arraystretch}{1.2}
\centering  
\caption{Traffic accident category in the DoTA dataset.}
\label{categories}
\resizebox{\linewidth}{!}{%
\begin{tabular}{cc}
\hline
Label         & Anomaly Category                    \\ \hline
ST & Collision with another vehicle that starts, stops, or is stationary \\
AH & Collision with another vehicle moving ahead or waiting \\ 
LA & Collision with another vehicle moving laterally in the same direction \\
OC & Collision with another oncoming vehicle \\ 
TC & Collision with another vehicle that turns into or crosses a road \\ 
VP & Collision between vehicle and pedestrian \\
VO & Collision with an obstacle in the roadway \\ 
OO & Out-of-control and leaving the roadway to the left or right \\ 
UK & Unknown \\ 
\hline
\end{tabular}}
\end{table}
\subsection{Dataset}
For the sake of fairness, we follow prior works \cite{9712446, 9714213} and evaluate our method on a recently publicly available dataset named DoTA \cite{9712446}. DoTA is the first traffic anomaly video dataset that provides detailed spatio–temporal annotations of anomalous objects for traffic accident detection in driving scenarios. It contains temporal, spatial, and categorical annotations of accidents for frames and objects in video. The DoTA consists of 4677 video clips with the resolution of $1280 \times 720$, the majority of which come from two YouTube channels that provide traffic accident videos for driver education purposes. It includes a variety of dashcam videos from different areas under different weather and lighting conditions.
Each video is annotated with an anomaly start and end time, which separates it into three parts: the precursor, which is the normal video that precedes the anomaly, the accident frames, and the post-accident frames. Moreover, each anomaly participant is labeled with a track ID, and their bounding box is labeled from anomaly start to anomaly end. Besides, each video is assigned to one of 9 categories, which we summarize in Table \ref{categories}.
\begin{table}[!t]
\renewcommand{\arraystretch}{1.1}
\centering  
\caption{The AUC $\uparrow$ (\%) of different approaches on the DoTA dataset.}
\label{AUC}
\begin{tabular}{ccc}
\hline
Methods         & Input                     & AUC  \\ \hline
ConvAE \cite{Hasan_2016_CVPR} & Gray                      & 64.3 \\
ConvAE \cite{Hasan_2016_CVPR} & Flow                      & 66.3 \\ 
ConvLSTMAE \cite{10.1007}     & Gray                      & 53.8 \\
ConvLSTMAE \cite{10.1007}     & Flow                      & 62.5 \\ 
AnoPred \cite{Liu_2018_CVPR}  & RGB                       & 67.5 \\ 
AnoPred \cite{Liu_2018_CVPR}  & Mask RGB                  & 64.8 \\
FOL-STD \cite{8967556}        & Box                       & 66.7 \\ 
FOL-STD \cite{8967556}        & Box + Flow                & 69.1 \\ 
FOL-STD \cite{8967556}        & Box + Flow + Ego          & 69.7 \\ 
FOL-Ensemble \cite{9712446}   & RGB + Box + Flow + Ego    & 73.0 \\ 
\bf{MAMTCF}                   & Box + Flow                & \bf{76.6} \\
\hline
\end{tabular}
\end{table}

In our experiments, we ensure fairness of comparison by adopting a data partitioning consistent with \cite{9712446, 9714213}, where DoTA is randomly divided into 3,275 training videos and 1,402 testing videos. During training, we only use the precursor frames from each video.

\subsection{Evaluation Setups}
\subsubsection{Metrics} Following prior works \cite{Luo_2017_ICCV, Gong_2019_ICCV, 9712446}, we use Area under ROC curve (AUC) metrics to evaluate the performance of different traffic accident detection models.

Area under ROC curve (AUC): Performance was evaluated by adopting the area under a standard frame-level receiver operating characteristic curve (ROC), with true positive rate (TPR) as the vertical axis and false positive rate (FPR) as the horizontal axis. The larger AUC prefers a better performance.
\subsubsection{Baselines}
To verify the superiority of the proposed framework, we compare with the following state-of-the-art methods.\\
\textbf{ConvAE} \cite{Hasan_2016_CVPR}: The spatio-temporal autoencoder-based model reconstructs the input and computes an anomaly score based on the reconstruction error. In the experiments, we compare two variants of ConvAE: one that reconstructs grayscale images and another that reconstructs optical flow.\\
\textbf{ConvLSTMAE} \cite{10.1007}: A method combines CNN and LSTM to model spatial and temporal features. In the experiments, we also compare two variants of reconstructing grayscale images and reconstructing optical flow.\\
\textbf{AnoPred} \cite{Liu_2018_CVPR}: A VAD method based on frame prediction takes the first four continuous RGB frames as input and applies UNet to predict a future RGB frame. In the experiments, we compare two variants: one that predicts the whole RGB frame and another that predicts only the RGB image of the foreground object. \\
\textbf{FOL-STD} \cite{8967556}: A traffic accident detection method is based on future object localization. In our experiments, we compare three variants: using only the bounding boxes of objects as input, using the bounding boxes of objects and the corresponding optical flow as input, and using the bounding boxes of objects, the optical flow, and the ego motion as input. \\
\textbf{FOL-Ensemble} \cite{9712446}: An ensemble approach. In this method, two models are trained separately, \textit{i.e.}, the method FOL-STD \cite{8967556} based on future object localization and the method AnoPred \cite{Liu_2018_CVPR} based on frame prediction. The anomaly score of each object is then mapped to per pixel score in each frame by a late fusion strategy. \\
\textbf{FOL}: Our approach without the optical flow reconstruction task and MAMR mechanism. \\
\textbf{FLOW}: Our approach without future object localization tasks and MAMR mechanism.
\begin{table*}[!t]
\renewcommand{\arraystretch}{1.1}
\centering 
\caption{The AUC $\uparrow$ (\%) of different methods for each individual accident class on the DoTA dataset is presented. The $*$ indicates non-ego accidents, while ego-involved accidents are shown without $*$. 
$-$ indicates that the method does not present AUC performance for the corresponding category of traffic accident. We \textbf{bold} the best performance and \underline{underline} the suboptimal.} 
\label{tab:my-table}
\begin{tabular}{cccccccccccc} \hline 
Methods & ST & AH & LA & OC & TC & VP & VO & OO & UK & AVG\\ \hline 
AnoPred \cite{Liu_2018_CVPR} & \underline{69.9} & 73.6 & \underline{75.2} & 69.7 & 73.5 & 66.3 & - & - & - & 71.4\\ 
AnoPred \cite{Liu_2018_CVPR}+Mask & 66.3 & 72.2 & 64.2 & 65.4 & 65.6 & \underline{66.6} & - & - & - & 66.7\\
FOL-STD \cite{8967556} & 67.3 & 77.4 & 71.1 & 68.6 & 69.2 & 65.1 & - & - & - & 69.7\\ 
FOL-Ensemble \cite{9712446}&\bf{73.3}&\bf{81.2}&74.0&\bf{73.4}&\underline{75.1}&\bf{70.1} & - & - & - & \bf{74.5}\\ 
\textbf{MAMTCF} & 64.0 & \underline{78.7} & \bf{82.5} & \underline{71.4} & \bf{77.6} & \underline{66.6} & \bf{76.8} & \bf{79.3} & \bf{71.8} & \underline{74.3}\\
\hline 
Methods & ST* & AH* & LA* & OC* & TC* & VP* & VO* & OO* & UK* & AVG\\ \hline 
AnoPred \cite{Liu_2018_CVPR} & 70.9 & 62.6 & 60.1 & 65.6 & 65.4 & 64.9 & 64.2 & 57.8 & - & 63.9 \\
AnoPred \cite{Liu_2018_CVPR}+Mask & 72.9 & 63.7 & 60.6 & 66.9 & 65.7 & 64.0 & 58.8 & 59.9 & - & 64.1 \\
FOL-STD \cite{8967556} & {75.1} & 66.2 & 66.8 & {74.1}&72.0&{69.7}&63.8&69.2 & - & 69.6 \\
FOL-Ensemble \cite{9712446}& \underline{77.5}&\underline{69.8}&\underline{68.1}&\underline{76.7}&\underline{73.9}&\underline{71.2}&\underline{65.2}&\underline{69.6} & - & \underline{71.5} \\
\textbf{MAMTCF} & \bf{78.8} & \bf{74.9} & \bf{76.1} & \bf{82.0} & \bf{76.8} & \bf{86.7} & \bf{77.2} & \bf{77.8} & \bf{76.7} & \bf{78.6} \\
\hline
\end{tabular} 
\end{table*}
\subsection{Quantitative Results}
\subsubsection{Overall results} We compare our proposed MAMTCF with all the above methods in terms of AUC metrics. Table \ref{AUC} summarizes the results of different algorithms and their corresponding inputs for each variant. Based on these results, we draw the following conclusions:
\begin{itemize}
    \item Overall, our method outperforms all the previous methods in terms of AUC.
    \item Our method outperforms traffic accident detection methods based on a single pretext task, \textit{i.e.}, appearance-based and future-object-localization-based TAD methods. For instance, compared to ConvAE \cite{Hasan_2016_CVPR} which is the reconstruction-based method and has an AUC of 66.3, our method (76.6) achieves a relative improvement of 15.5\% on AUC. Similarly, compared to AnPred \cite{Liu_2018_CVPR} which is a prediction-based method and has an AUC of 67.5, our method achieves a relative improvement of 13.5\%. Besides, our method achieves a relative improvement of 9.9\% compared to the previous best future-object-localization-based method FOL-STD \cite{8967556} (69.7). Furthermore, our method also outperforms the ensemble model FOL-Ensemble \cite{9712446}, which is the previous best method, by achieving a relative improvement of 4.9\%. This demonstrate that our proposed multi-task collaborative framework to model both the appearance changes of video frames and the motion states of objects in the scene is indeed helpful for TAD in driving videos.
    \item In addition, our method achieves a 10.9\% relative improvement compared to FOL-STD \cite{8967556} under the same input conditions, \textit{i.e.}, the extracted bounding box of the object and the optical flow of the video frame (Box + Flow). This indicates that our proposed multi-task collaborative framework can make better use of data and is more suitable for traffic accident detection tasks.
\end{itemize}
\subsubsection{Per-class results} 
To investigate the detection capability of our proposed framework for ego-involved and non-ego traffic accidents, we further compare the detection performance of different methods for these two types of accidents. Table \ref{tab:my-table} summarizes the detection AUC performance of different approaches for 9 categories of traffic accidents on the DoTA dataset. We present results for ego-involved accidents and non-ego traffic accidents (marked by $*$) separately. Additionally, we report the average performance of each method in the last column. In general, our method outperforms previous single-pretext-task-based methods in terms of average AUC for both ego-involved and non-ego accidents. For example, in non-ego accident detection, our method improves the average AUC by 12.9\% compared to the previous best single-pretext-task-based method, FOL-STD. In ego-involved accident detection, our approach achieves a relative improvement of 6.6\% compare to previous best future-object-localization-based method, FOL-STD, and a relative improvement of 4.1\% compare to previous prediction-based method, AnPred. These results indicate that the proposed multi-task collaborative framework can indeed promote the detection of both ego-involved and non-ego traffic accidents.
Notably, our method also greatly outperforms FOL-Ensemble in terms of the average AUC of non-ego accident detection. Although FOL-Ensemble achieves comparable AUC performance on average to our method on ego-involved accident detection, it integrates two separately trained single-pretext-task-based TAD  methods using a late fusion strategy. Such ensemble models undoubtedly increase the difficulty of traffic accident detection and potentially limit its application in autonomous driving and driver assistance systems.
Besides, our method has better detection capacity in most accident categories, especially non-ego accidents. 
The main reason for this is that our proposed MAMR mechanism fully explores the interrelation between different types of motion representations and augments the motion representations by exploiting the high-level features of normal traffic patterns stored in memory, thus increasing the difference from anomalies.
\begin{table}[!t]
\renewcommand{\arraystretch}{1.1}
\centering  
\caption{The AUC $\uparrow$ (\%) scores for variants on the DoTA dataset.}
\label{ablation}
\begin{tabular}{cc|ccc|c}
\hline
\multicolumn{2}{c|}{Pretext Task}                   & \multicolumn{3}{c|}{Fusion Strategy}                       & \multicolumn{1}{c}{\multirow{2}{*}{AUC}} \\ \cline{1-5}
FOL & FLOW & Concat & Transformer & MAMR &                     \\ \hline
\checkmark & $\times$     & $-$ & $-$ & $-$ & 62.6 \\
$\times$   & \checkmark & $-$ & $-$ & $-$ & 71.3 \\
\checkmark & \checkmark & \checkmark & $-$ & $-$ & 75.8 \\
\checkmark & \checkmark & $-$ & \checkmark & $-$ & 76.1 \\
\checkmark & \checkmark & $-$ & $-$ & \checkmark & \textbf{76.6}\\
\hline
\end{tabular}
\end{table}
\begin{figure*}[!t]
	\centering
	\includegraphics[width=0.97\linewidth]{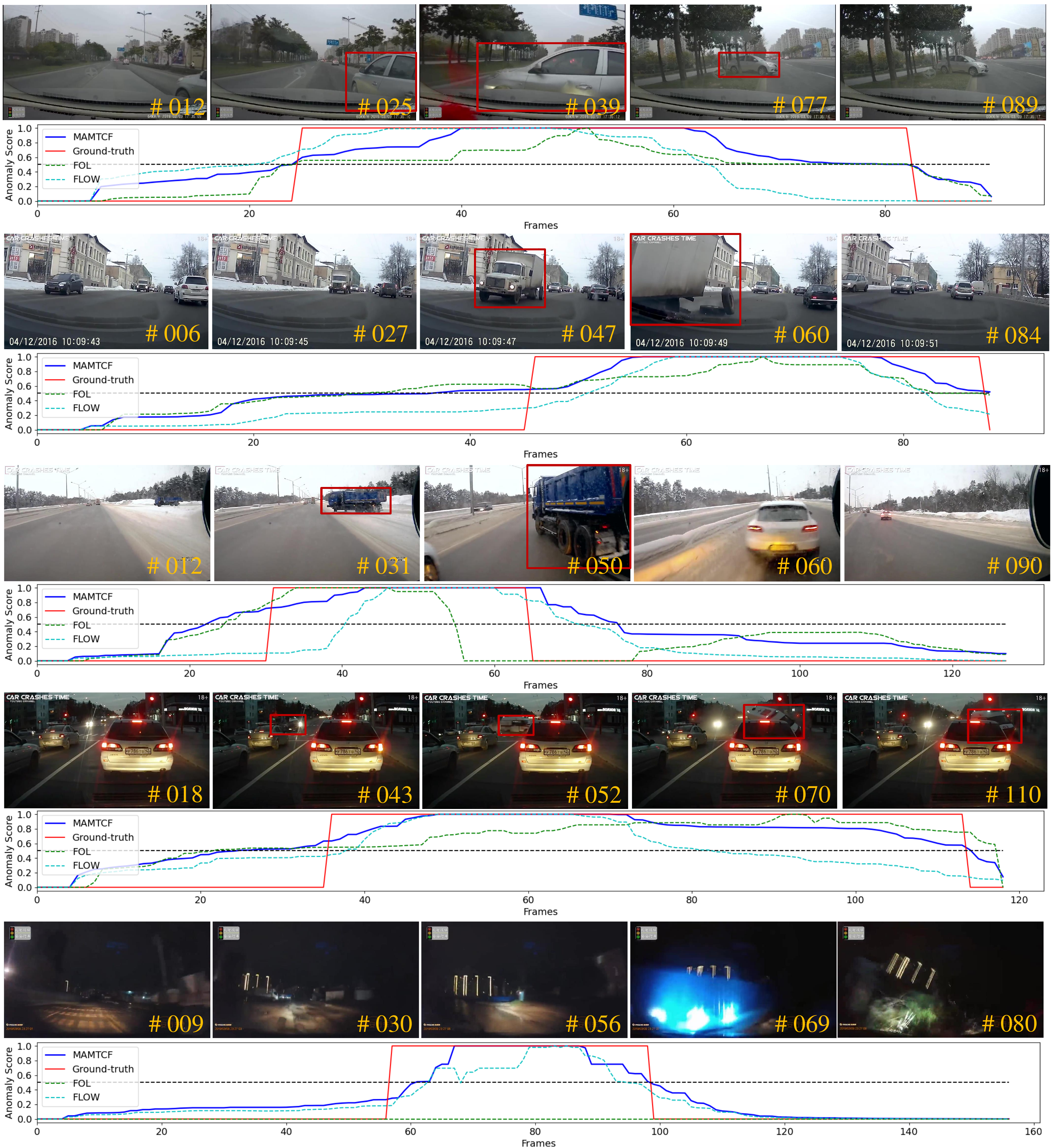}
	\caption{The visualization of anomaly score curves for traffic accident detection of different variants on the DoTA dataset. The first row of each case shows the extracted video frames of the driving video, where the red boxes mark the object involved in the accident. The second rows show the anomaly score curves of different methods on the corresponding whole videos. Better viewed in color.}
	\label{fig:vis}
\end{figure*}
\subsection{Qualitative Results}
In this subsection, we provide visual examples in Fig. \ref{fig:vis} to illustrate that our multi-task collaborative framework MAMTCF can better detect ego-involved and non-ego traffic accidents than single-pretext-task-based TAD methods.

Overall, as shown in Fig. \ref{fig:vis}, our method performs better than the TAD methods based on a single pretext task, \textit{i.e.}, future object localization (FOL) and optical flow reconstruction (FLOW) tasks, when comparing the anomaly score curves of the different methods with the ground truth curves. Specifically, we show five traffic accident types as examples from top to bottom:
a) The ego-vehicle collides with another vehicle moving laterally in the same direction.
b) The ego-vehicle collides with an abnormal vehicle traveling in the opposite direction.
c) The ego-vehicle collides with another vehicle turning into the road.
d) The other vehicle collides with another vehicle crossing the road.
e) The ego-vehicle is out of control.
From the above visualization results of different types of traffic accidents, we can summarize as follows. First, the FOL-based TAD method focus on detecting anomalies in the motion trajectory of observed objects. However, these types of approaches may not be able to detect ego-involved anomalies when there are no objects in the scene (as shown in the last row of Fig. \ref{fig:vis}). Second, the FLOW-based method concentrate on detecting changes in appearance of video frames but may cause false detection or missed detection of traffic accidents due to the rapid movement of dashboard-mounted cameras (as shown in 5 traffic accident types in Fig. \ref{fig:vis}). Finally, our memory-augmented multi-task collaborative framework can absorb the advantages of these two types of single pretext tasks based methods to detect both ego-involved and non-ego traffic accidents.
\subsection{Ablation Investigation}
In this subsection, we conduct ablation experiments to investigate how our proposed MAMTCF framework impacts the detection of traffic accident.
\subsubsection{Variants of our architecture}
We evaluate each component of the proposed framework by performing a series of ablation experiments, including pretext tasks (\textit{i.e.}, FOL and FLOW tasks) and fusion strategies for collaborating multi-tasks. The experimental results are summarized in Table \ref{ablation}. Note that in our variant, the FOL-based TAD method uses $S_l$ in formula (\ref{formula}) to calculate the final anomaly scores, while the FLOW-based method uses $S_e$ to compute the final anomaly scores. From the results reported in Table \ref{ablation}, we have the following conclusions. First, the multi-task based methods significantly outperform those based on a single pretext task, namely FOL and FLOW. The main reason is that the FOL-based method focuses on the motion states of observed objects and helps detect non-ego accidents, while the FLOW-based method focuses on appearance changes and helps detect ego-involved accidents. Therefore, combining the two pretext tasks promotes to detect both types of traffic accidents. Second, modeling the interrelation between \textit{global motion} and \textit{object motion} is beneficial for modeling both appearance changes in video frames and motion states of observed objects. By using Transformers instead of simple concatenation, we can observe further improvements in the performance of traffic accident detection. Third, the best fusion strategy for collaborating multi-tasks is our proposed MAMR mechanism. 
This is because we further augment the motion representation with high-level features of normal traffic patterns stored in memory, thus increasing the difference from anomalies.
\begin{figure}[!t]
	\centering
	\includegraphics[width=1.0\linewidth]{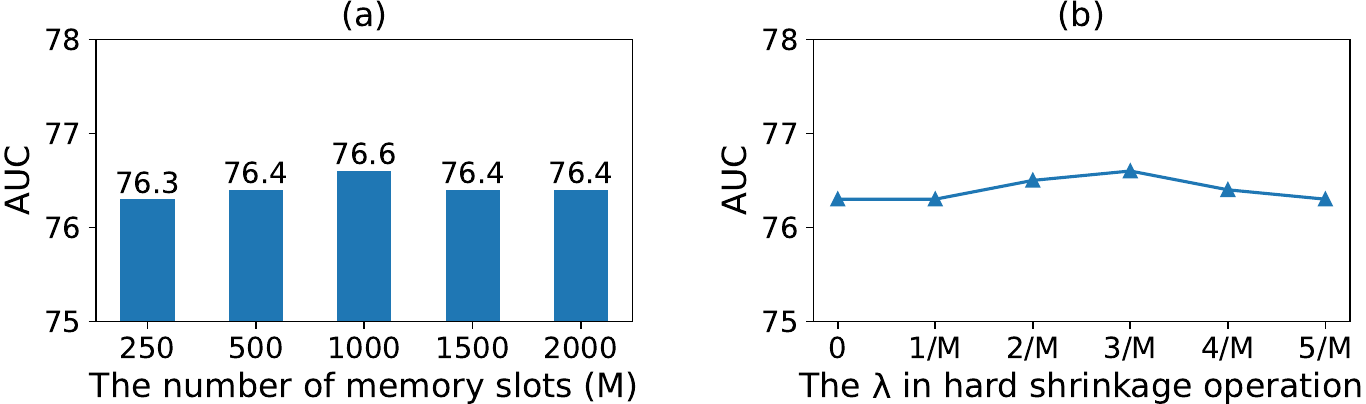}
	\caption{Hyperparameters experiments on memory in MAMR mechanism were conducted. Fig. (a) shows the ablation experiment with the number of memory slots $M$ for the hard shrinkage operation threshold $\lambda = 3/M$, and Fig. (b) shows the ablation experiment with the threshold $\lambda$ for memory slots $M=1000$.}
	\label{fig:result}
\end{figure}
\begin{table}[!t]
\renewcommand{\arraystretch}{1.1}
\centering  
\caption{Ablation experiments on the layer $L$ of the MAMR mechanism.}
\label{transformer block}
\begin{tabular}{cc}
\hline
Number of layers ($L$) & AUC  \\ \hline
1                      & 75.9 \\
2                      & 76.0 \\
3                      & \bf{76.6} \\ 
4                      & 76.3 \\
5                      & 76.2 \\ 
\hline
\end{tabular}
\end{table}
\subsubsection{Hyperparameters analysis}
To ensure the rationality of our approach’s hyperparameter settings, we further conduct ablation experiments on the number $M$ of memory slots, the threshold $\lambda$ for hard shrinkage operations, and the layer $L$ of MAMR mechanism. Fig. \ref{fig:result} illustrates the results of hyperparameter experiments for memory in the MAMR mechanism. To investigate the robustness of the proposed method to memory size $M$, we conduct experiments with different numbers of memory slots, and the corresponding AUC performances are shown in Fig. (a). From the results, we can conclude that our method is robust to changes in memory slot size and achieves good accident detection performance even with small memory sizes. With an increase in the number of memory slots, the performance of accident detection improves slightly, and best performance is achieved when $M=1000$. However, when the number of slots continues to increase, the performance of accident detection slightly decreases. One reason for this is that the memory size is too large, which may causes the model to overfit. Fig. (b) shows the effect of the hard shrinkage operation on the MAMR mechanism at different threshold settings. Note that when the threshold $\lambda=0$, we apply the softmax operation instead of the hard shrinkage operation. Fig. (b) shows that the hard shrinkage operation performs slightly better than the softmax operation because it improves the sparsity of the memory, thus encouraging to reconstruct motion representations with fewer but more relevant memory items. Additionally, the best accident detection performance is achieved when $\lambda=3/M$. When $\lambda>3/M$, the memory items may be too sparse and result in a slight drop in performance. In addition, we conduct experiments on the layer $L$ of MAMR mechanism to explore their impact on traffic accident detection. The experimental results are summarized in Table \ref{transformer block}. One observation from these experiments is that increasing the layer of MAMR mechanism not only increases the parameters of the model but may also compromise the performance of accident detection. Based on the experimental results, we finally adopted $L=3$ in our experiments.
\begin{figure*}[!t]
	\centering
	\includegraphics[width=0.97\linewidth]{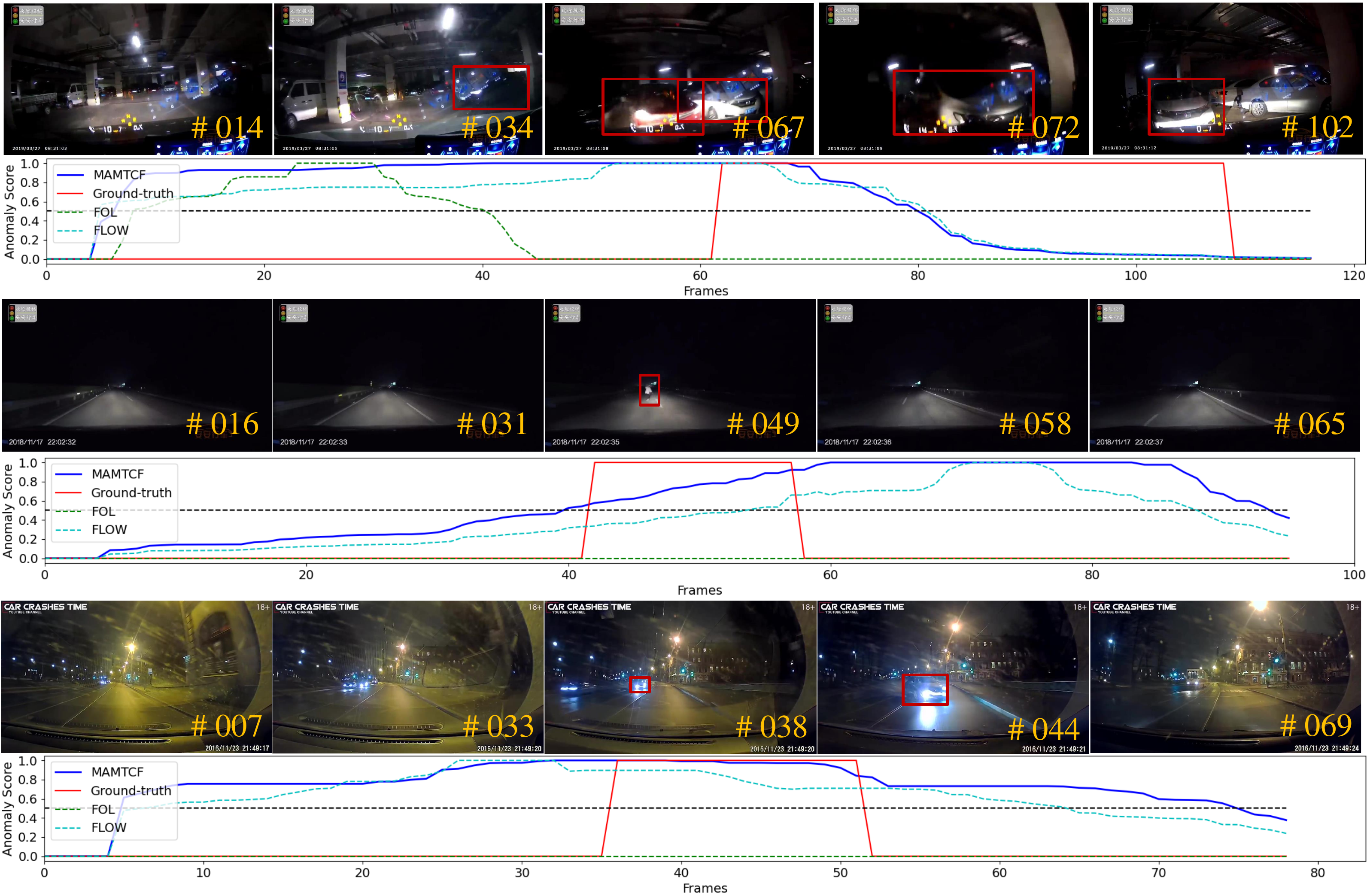}
	\caption{Visualization of some failure cases of the proposed MAMTCF. Better viewed in color.}
	\label{fig:dis}
\end{figure*}
\subsection{Disscusion}
In this subsection, we discuss some limitations of our approach. First, we experimentally find that our method may fail to accurately detect traffic accidents in scenarios with small changes in the motion of ego-vehicle. Fig. \ref{fig:dis} shows several cases where our approach fails. Specifically, the first scenario shows the ego-vehicle colliding with a stationary vehicle and coming to a stop. The second scenario shows the ego-vehicle colliding with a pedestrian on the road and continuing to move. The third scenario shows the ego-vehicle colliding with an oncoming vehicle and then coming to a stop. Analyzing the anomaly score curve in Fig \ref{fig:dis}, we can attribute the main reason why our method fails in the above scenario to two factors. On the one hand, object detection algorithms may fail in some scenarios (\textit{e.g.}, pedestrians appearing suddenly in the dark or bright light from the oncoming vehicle), which can invalidate future object localization algorithms. On the other hand, due to the slow movement of the vehicle or complex illumination in the dark, the optical flow of video frames changes little (\textit{i.e.}, the appearance changes are not obvious), which leads to the failure of TAD method based on optical flow reconstruction to detect traffic accidents. This also explains the poor performance of our method on detecting ego-involved ST, OC, and VP accident categories in Table \ref{tab:my-table}. Although our method outperforms the single-pretext-task-based TAD methods in most scenarios, it is clear that more efforts are needed by the community in the future to explore a general traffic accident detection method for the above scenarios.

In addition, most existing unsupervised traffic accident detection methods (including ours) are based on a two-stage strategy. In the first stage, trained models are used to extract features such as optical flow or bounding boxes of objects. The second stage then trains the pretext task for traffic accident detection. This two-stage approach not only increases the time required for accident detection but also suffers from the error accumulation problem, which limits the application of traffic accident detection algorithms. Therefore, in the future, we will focus on exploring one-stage unsupervised traffic accident detection methods.

\section{Conclusion}  \label{sec:conclusion}
In this paper, we have proposed a novel memory-augmented multi-task collaborative framework for unsupervised traffic accident detection in driving videos. Different from previous TAD methods based on a single pretext task, our method collaborates optical flow reconstruction with future object localization tasks to better detect both ego-involved and non-ego accidents. Furthermore, 
the proposed MAMR mechanism fully explores the interrelation between different types of motion representations, and augments the motion representations with high-level features of normal traffic patterns stored in memory, thus broadening the distinction from anomalies.
Both quantitative and qualitative experimental results on recently published large-scale dataset have demonstrated the superiority of our approach in various situations.

\ifCLASSOPTIONcaptionsoff
\newpage
\fi

\bibliography{MTCF}
\bibliographystyle{IEEEtran}
\end{document}